%% file: ScoreFollowingGame.tex
\documentclass{article}
\usepackage{ismir,amsmath,cite,url}
\usepackage{graphicx}
\usepackage{amssymb}

\usepackage{booktabs}       
\usepackage{multirow}

\usepackage{marvosym}

\usepackage[dvipsnames]{xcolor}

\newcommand{\baseline}{MM-Loc }

\title{Learning to Listen, Read, and Follow: \\
Score Following as a Reinforcement Learning Game}

\multauthor
{Matthias Dorfer$^{*}$ \hspace{1cm} Florian Henkel$^{*}$ \hspace{1cm} Gerhard Widmer$^{*\dagger}$}
{$^*$Institute of Computational Perception, Johannes Kepler University Linz, Austria\\
$^\dagger$The Austrian Research Institute for Artificial Intelligence (OFAI), Austria\\
{\tt matthias.dorfer@jku.at}
}
\def\authorname{Matthias Dorfer, Florian Henkel, Gerhard Widmer}

\begin{document}

\maketitle
\begin{abstract}
\input{abstract}
\end{abstract}

\section{Introduction}
\label{sec:introduction}
\input{introduction}

\section{Description of Data}
\label{sec:data}
\input{data}

\section{Score Following as a\\ Markov Decision Process}
\label{sec:mdp}
\input{mdp}

\section{Learning to Follow}
\label{sec:rl}
\input{rl}

\section{Experimental Results}
\label{sec:experimental_results}
\input{experiments}

\section{Discussion and Conclusion}
\label{sec:conclusion}
\input{conclusion}

\section{Acknowledgements}
This work is supported by the European Research Council (ERC Grant Agreement
670035, project CON ESPRESSIONE).

\bibliography{ISMIRtemplate}

%
%
%
%

\end{document}

%% file: abstract.tex
Score following is the process of tracking a
musical performance (audio) with respect to
a known symbolic representation (a score).
We start this paper by formulating score following
as a multimodal Markov Decision Process,
the mathematical foundation for sequential decision making.
Given this formal definition, we address the score following task with state-of-the-art
deep reinforcement learning (RL) algorithms
such as synchronous advantage actor critic (A2C).
In particular, we design multimodal RL agents
that simultaneously learn to listen to music,
read the scores from images of sheet music,
and follow the audio along in the sheet, in an end-to-end fashion.
All this behavior is learned entirely from scratch,
based on a weak and potentially delayed reward signal
that indicates to the agent how close it is to the correct position in the score.
Besides discussing the theoretical advantages of this learning paradigm,
we show in experiments
that it is in fact superior compared to previously proposed methods for score following in raw sheet music images.

%% file: introduction.tex
\begin{figure}[t!]
 \centerline{\includegraphics[width=0.85\columnwidth]{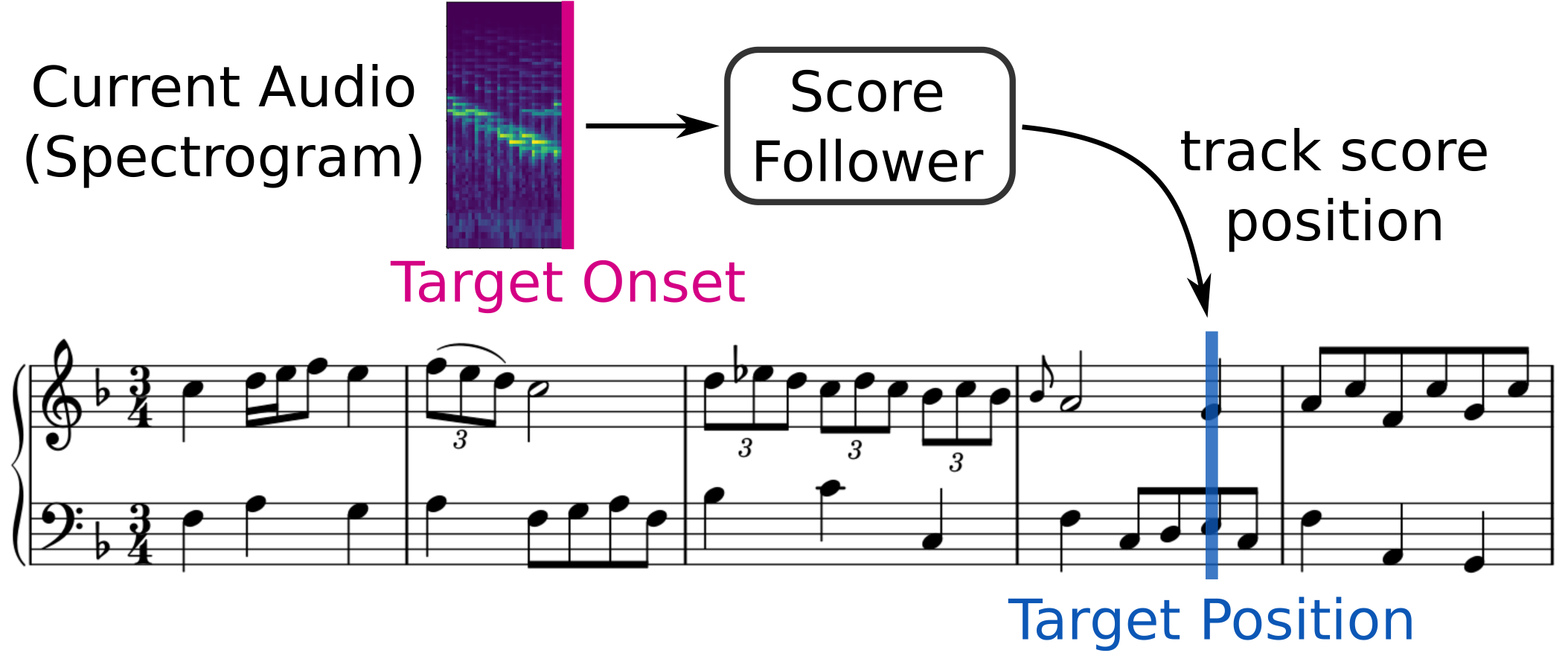}}
 \caption{Sketch of score following in sheet music.
 Given the incoming audio, the score follower has to track the corresponding position in the score (image).
}
\label{fig:sf_sketch}
\end{figure}

This paper addresses the problem of score following in sheet music images.
The task of an automatic score following system is to follow a musical performance
with respect to a known symbolical representation, the score (cf. Figure \ref{fig:sf_sketch}).
In contrast to audio-score alignment in general \cite{muller:book:2015},
all of this takes place in an on-line fashion.
Score following itself has a long history in Music Information Retrieval (MIR)
and forms the basis for many subsequent applications such as
automatic page turning \cite{Arzt_2008_PageTurning},
automatic accompaniment \cite{Cont_2009_CDFA,Raphael_2010_MusicPlusOne}
or the synchronization of visualizations to the live music during concerts \cite{Arzt_2015_AIConcertgebouw,Prockup_2013_Orchestra}.

Traditional approaches to the task depend on a symbolic,
computer-readable representation of the score,
such as MusicXML or MIDI (see e.g. \cite{kurth2007automated,Nakamura_2015_ScoreFollowing,Arzt_2015_AIConcertgebouw,Prockup_2013_Orchestra,Cont_2009_CDFA,Raphael_2010_MusicPlusOne,Miron_2014_Alignment,Duan_2011_Alignment,izmirli2012bridging}).
This representation is created either manually
(e.g. via the time-consuming process of (re-)setting the score in a music notation program),
or automatically via optical music recognition software \cite{HaPe2017MUSCIMADataset,byrd:jnmr:2015,BalkePM15_MatchingMusicalThemes_ICASSP}.
However, automatic methods are still
unreliable and thus of limited use,
especially for more complex music like orchestra pieces \cite{Thomas_2012_LinkingAudioAndSheetMusic}.

To avoid these complications, 
\cite{Dorfer2016Towards}
proposes a multimodal deep neural network that
directly learns to match sheet music and audio in an end-to-end fashion.
Given short excerpts of audio and the corresponding sheet music, the network learns to predict
which location in the given sheet image best matches the current audio excerpt.
In this setup, score following can be formulated as a multimodal localization task.
However, one problem with this approach is that
successive time steps are treated independently from each other.
We will see in our experiments that this causes jumps in the tracking process
especially in the presence of repetitive passages.
A related approach \cite{Dorfer2017Audio2Score} trains
a multimodal neural network to learn
a joint embedding space for
snippets of sheet music and corresponding short excerpts of audio.
The learned embedding allows to compare observations across modalities,
e.g., via their cosine distance.
This learned cross-modal similarity measure
is then used to compute an off-line alignment
between audio and sheet music via dynamic time warping.

Our proposal is inspired by these works,
but uses a fundamentally different machine learning paradigm.
The central idea is to interpret score following as a \emph{multimodal control problem} \cite{Duan_2016_RLControl}
where the agent has to navigate through the score by adopting its reading speed
in reaction to the currently playing performance.
To operationalize this notion, we formulate score following as a \emph{Markov Decision Process (MDP)} in Section \ref{sec:mdp}.
MDPs are the mathematical foundation for sequential decision making
and permit us to address the problem with state-of-the-art \emph{Deep Reinforcement Learning (RL)} algorithms (Section \ref{sec:rl}).
Based on the MDP formulation, we design agents that consider both
the score and the currently playing music to achieve an overall goal,
that is to track the correct position in the score for as long as possible.
This kind of interaction is very similar to controlling an agent in a video game,
which is why we term our MDP the \emph{score following game};
it is in fact inspired by the seminal paper by Mnih et al. \cite{Mnih_2015_DeepRL}
which made a major contribution to the revival of deep RL by
achieving impressive results in a large variety of Atari games.
In experiments with monophonic as well as polyphonic music (Section \ref{sec:experimental_results}),
we will show that the RL approach is indeed
superior to previously proposed score following methods \cite{Dorfer2016Towards}.
The code for the score following game is available at \url{https://github.com/CPJKU/score_following_game}.

%% file: data.tex
To set the stage, we first need to describe the kind of data needed for training and evaluating
the multimodal RL score following agents.
We assume here that we are given a collection of piano pieces
represented as pairs of audio recordings and sheet music images.
In order to train our models and to later quantify the score following error,
we first need to establish correspondences between
individual pixel locations of the note heads in a sheet and
their respective counterparts (note onset events)
in the respective audio recordings.
This has to be done either in a manual annotation process
or by relying on synthetic training data which is generated from digital sheet music formats
such as \emph{Musescore} or \emph{Lilypond}.
As this kind of data representation is identical to the one used in \cite{Dorfer2016Towards,Dorfer2017Audio2Score}
we refer to these works for a detailed description of the entire alignment process.

%% file: mdp.tex
\begin{figure}[ht!]
 \centerline{\includegraphics[width=0.95\columnwidth]{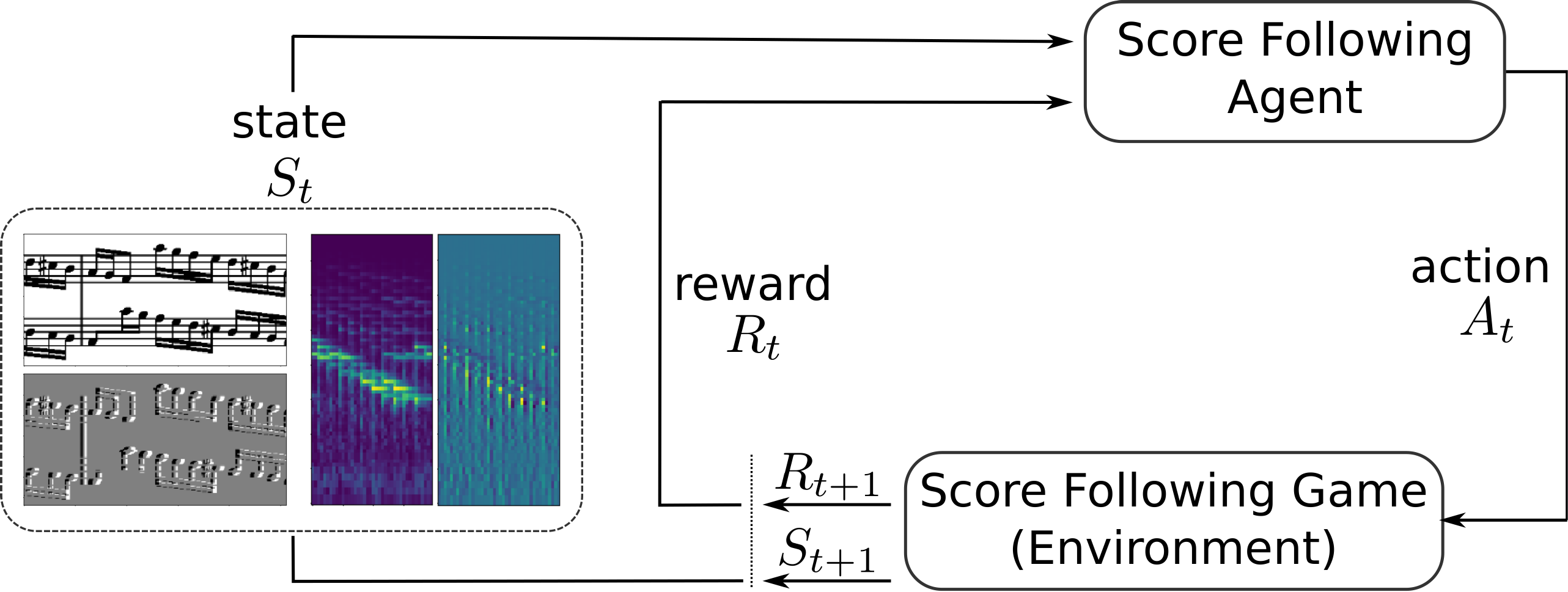}}
 \caption{Sketch of the score following MDP. The agent receives the current state of the environment $S_t$ and a scalar reward signal $R_t$ for the action taken in the previous time step. Based on the current state it has to choose an action (e.g. decide whether to increase, keep or decrease its speed in the score) in order to maximize future reward by correctly following the performance in the score.
}
\label{fig:mdp}
\end{figure}

Reinforcement learning can be seen as a computational approach to learning from interaction to achieve a certain predefined goal.
In this section, we formulate the task of score following as a \emph{Markov Decision Process (MDP)},
the mathematical foundation for reinforcement learning
or, more generally,
for the problem of sequential decision making\footnote{The notation in this paper follows the book by Barto and Sutton \cite{SuttonB_1998_RL}}.
Figure \ref{fig:mdp} provides an overview of the components involved in the score following MDP.

The \emph{score following agent} (or \emph{learner}) is the active component
that interacts with its \emph{environment},
which in our case is the score following game.
The interaction takes place in a closed loop where
the environment confronts the agent with a new situation (a \emph{state $ S_t$})
and the agent has to respond by making a decision,
selecting one out of a predefined set of possible \emph{actions} $A_t$.
After each action taken the agent receives the next state $S_{t+1}$
and a numerical \emph{reward signal} $R_{t+1}$
indicating how well it is doing in achieving the overall goal.
Informally, the agent's goal in our case is to track a performance in the score as accurately and robustly as possible; this criterion will be formalized in terms of an appropriate reward signal in Section \ref{subsec:reward} below.
By running the MDP interaction loop we end up with a sequence of states, actions and rewards
$S_0, A_0, R_1, S_1, A_1, R_2, S_2, A_2, R_3, ...$,
which is the kind of experience a RL agent is learning its behavior from.
We will elaborate on different variants of the learning process in Section \ref{sec:rl}.
The remainder of this section specifies all components of the score following MDP in detail.
In practice, our MDP is implemented as an environment in OpenAI-Gym\footnote{\url{https://gym.openai.com/}},
an open source toolkit for developing and comparing reinforcement learning algorithms.

\subsection{Score Following Markov States}
\label{subsec:states}
\begin{figure}[t!]
 \centerline{\includegraphics[width=0.95\columnwidth]{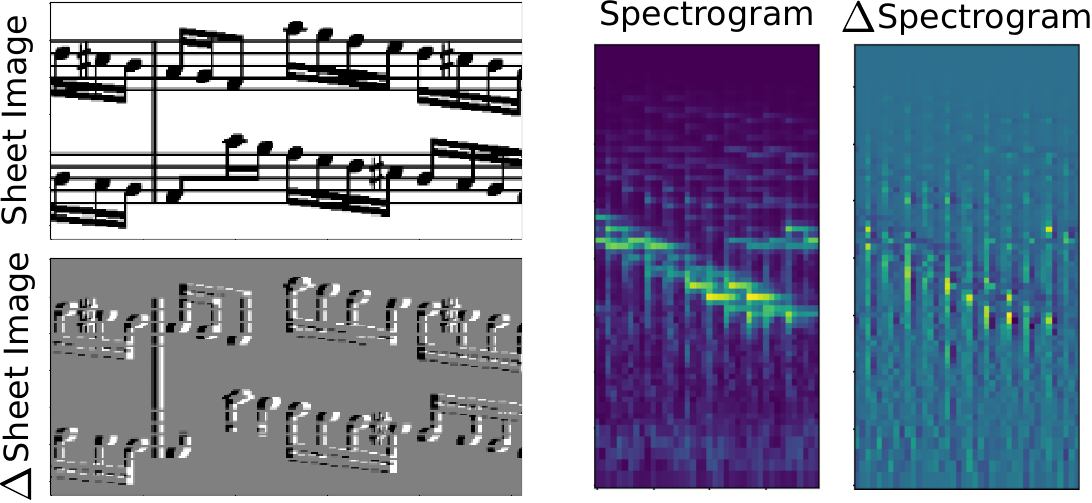}}
 \caption{Markov state of the score following MDP: the current sheet sliding window and spectrogram excerpt.
 To capture the dynamics of the environment we also add the one step differences ($\Delta$) wrt. the previous time step (state).
}
\label{fig:mdp_state}
\end{figure}

Our agents need to operate on two different inputs at the same time,
which together form the state $S_t$ of the MDP:
input modality one is a sliding window of the sheet image of the current piece,
and modality two is an audio spectrogram excerpt of the most recently played music ($\sim$ 2 seconds).
Figure~\ref{fig:mdp_state} shows an example of this input data for a piece by J.S. Bach.
Given the audio excerpt as an input the agent's task is to navigate through the global score
to constantly receive sheet windows from the environment
that match the currently playing music.
How this interaction with the score takes place is explained in the next subsection.
The important part for now is to note that score following embodies
dynamics which have to be captured by our state formulation,
in order for the process to satisfy the Markov property.
Therefore, we extend the state representation by adding the one step differences ($\Delta$)
of both the score and the spectrogram.
With the $\Delta$ images and spectrograms a state contains all the information needed by the agent
to determine where and how fast it is moving along in the sheet image.

\subsection{Agents, Actions and Policies}
The next item in the MDP (Figure \ref{fig:mdp}) is the agent, which is
the component interacting with the environment by taking actions
as a response to states received.
As already mentioned, we interpret score following as a multimodal control problem
where the agent decides how fast it would like to progress in the score.
In more precise terms, the agent controls its score progression speed $v_{pxl}$ in \emph{pixels per time step}
by selecting from a set of actions $A_t \in \{ -\Delta v_{pxl}, 0, +\Delta v_{pxl} \}$
after receiving state $S_t$ in each time step.
Actions $\pm \Delta v_{pxl}$ increase or decrease the speed
by a value of $\Delta v_{pxl}$ pixels per time step.
Action $a_1=0$ keeps it unchanged.
To give an example: a pixel speed of $v_{pxl}=14$ would shift the sliding sheet window
14 pixels forward (to the right) in the global unrolled score.

Finally, we introduce the concept of a \emph{policy} $\pi_{\Theta}(a|s)$ to define an agent's behavior.
$\pi$ is a conditional probability distribution
over actions conditioned on the current state.
Given a state $s$, it computes an action selection probability $\pi_{\Theta}(a|s)$
for each of the candidate actions $a \in A_t$.
The probabilities are then used for sampling one
of the possible actions.
In Section \ref{sec:rl} we explain how to use deep neural networks
as function approximators for policy $\pi_{\Theta}$ by optimizing the parameters $\Theta$ of a policy network.

\subsection{Goal Definition: Reward Signal and State Values}
\label{subsec:reward}
%
In order to learn a useful action selection policy, the agent needs \emph{feedback}. This means that we need to define
how to report back to the agent how well it does in accomplishing the task
and, more importantly, what the task actually is.

The one component in an MDP that defines the overall goal
is the reward signal $R_t \in \mathbb{R}$.
It is provided by the environment in form of a scalar,
each time the agent performs an action.
The \emph{sole objective of a RL agent is to maximize the cumulative reward over time}.
Note, that achieving this objective requires foresight and planning,
as actions leading to high instantaneous reward might lead to unfavorable situations in the future.
To quantify this longterm success,
RL introduces the \emph{return} $G$ which is defined as the discounted cumulative future reward: $G_t=R_{t+1} + \gamma R_{t+2} + \gamma^2 R_{t+3} + \cdots$.
The discount rate $\gamma$ (with $0.0 < \gamma \leq 1.0$, in our case $0.9$) is a hyper-parameter
assigning less weight to future rewards if smaller than $1.0$.

\begin{figure}[t!]
 \centerline{\includegraphics[width=0.95\columnwidth]{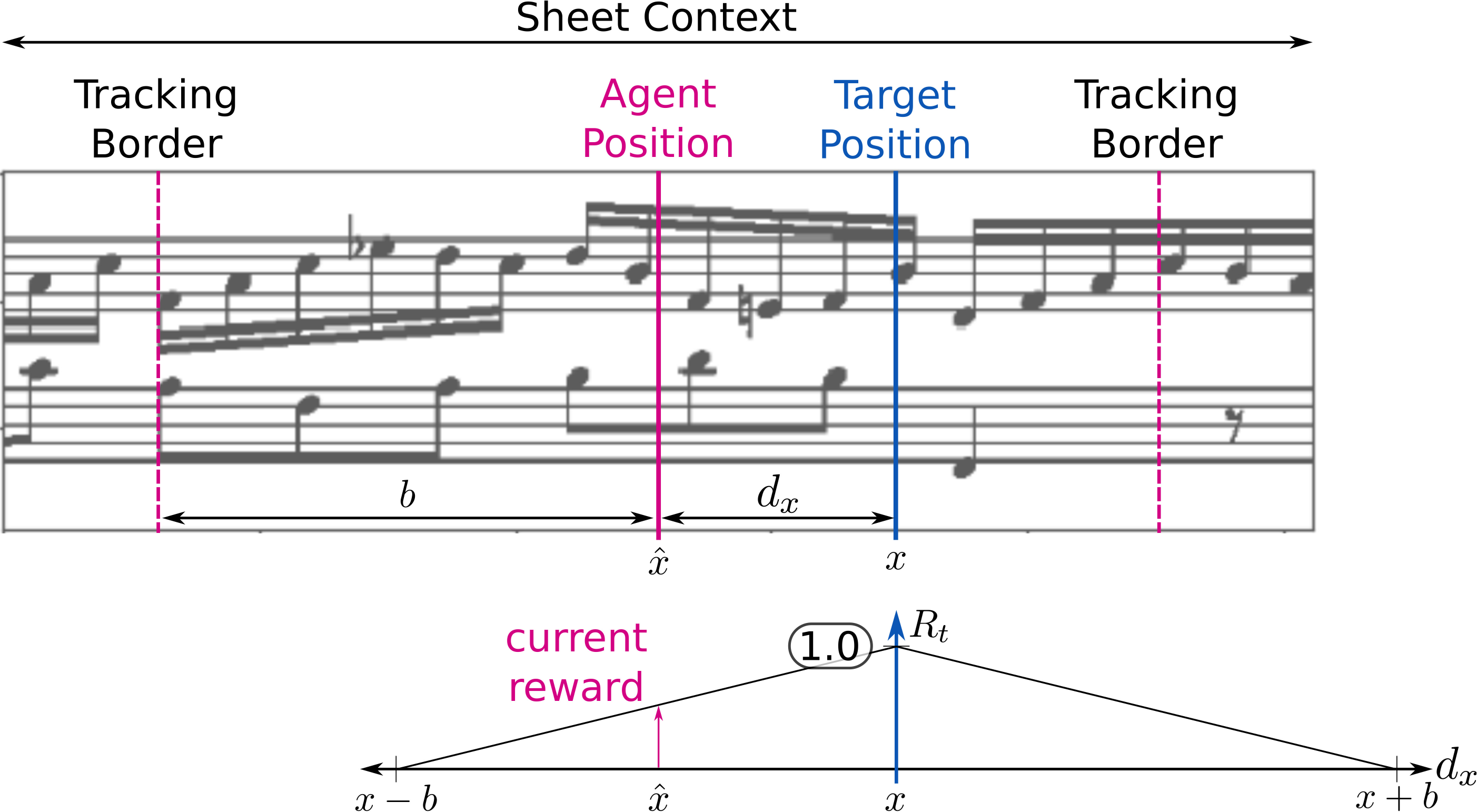}}
 \caption{Reward definition in the score following MDP.
 The reward $R_t$ decays linearly (range [0, 1]) depending on the agent's distance $d_x$ to the current true score position $x$.
}
\label{fig:reward}
\end{figure}

Figure \ref{fig:reward} summarizes the reward computation in our score following MDP.
Given annotated training data as described in Section \ref{sec:data},
the environment knows, for each onset time in the audio,
the true target position $x$ in the score.
From this, and the current position $\hat{x}$ of the agent, we
compute the current tracking error as $d_x=\hat{x} - x$, and define the reward signal $r$ within a predefined \emph{tracking window} $[x-b, x+b]$
around target position $x$ as: $r = 1.0 - |d_x| / b$.
Thus, the reward per time step reaches its maximum of $1.0$
when the agent's position is identical to the target position,
and decays linearly towards $0.0$ as the tracking error reaches the maximum permitted value $b$ given by the window size.
Whenever the absolute tracking error exceeds $b$ (the agent drops out of the window),
we reset the score following game (back to start of score, first audio frame).
As an RL agent's sole objective is to maximize cumulative future reward,
it will learn to match the correct position in the score and
to not lose its target by dropping out of the window.
We define the target onset,
corresponding to the target position in the score,
as the \emph{rightmost frame} in the spectrogram excerpt.
This allows to run the agents on-line, introducing only the delay
required to compute the most recent spectrogram frame.
In practice, we linearly interpolate the score positions for spectrogram frames
between two subsequent onsets in order to produce a continuous and stronger learning signal
for training.

As with policy $\pi$, we will use function approximation
to predict the future cumulative reward for a given state $s$,
estimating how good the current state actually is.
This estimated future reward is termed the \emph{value} $V(s)$ of state $s$.
We will see in the next section how state-of-the-art RL algorithms use these value estimates
to stabilize the variance-prone process of policy learning.

%% file: rl.tex
Given the formal definition of score following as an MDP
we now describe how to address it
with reinforcement learning.
Note that there is a large variety of RL algorithms.
We focus on \emph{policy gradient methods},
in particular the class of \emph{actor-critic methods}, due to their reported success in solving control problems \cite{Duan_2016_RLControl}.
The learners utilized are \emph{REINFORCE with Baseline} \cite{Williams_1992_Reinforce} and \emph{Synchronous Advantage Actor Critic (A2C)} \cite{Mnih_2016_A3C,Wu_2017_ScalableTRM},
where the latter is considered a state-of-the-art approach.
As describing the methods in full detail
is beyond the scope of this paper,
we provide an intuition on how the methods work
and refer the reader to the respective papers.

\subsection{Policy and State-Value Approximation via DNNs}
\label{subsec:function_approx}
In Section \ref{sec:mdp}, we introduced \emph{policy} $\pi_{\Theta}$,
determining the behavior of an agent,
and \emph{value function} $V(s)$,
predicting how good a certain state $s$ is with respect to cumulative future reward.
Actor-critic methods make use of both concepts.
The actor is represented by policy $\pi_{\Theta}$ and is responsible for selecting the appropriate action in each state.
The critic is represented by the value function $V(s)$ and helps the agent to judge how good the selected actions actually are.
In the context of deep RL both functions are approximated via a Deep Neural Network (DNN),
termed policy and value network.
We denote the parameters of the policy network with $\Theta$ in the following.

\begin{figure}[t!]
 \centerline{\includegraphics[width=0.85\columnwidth]{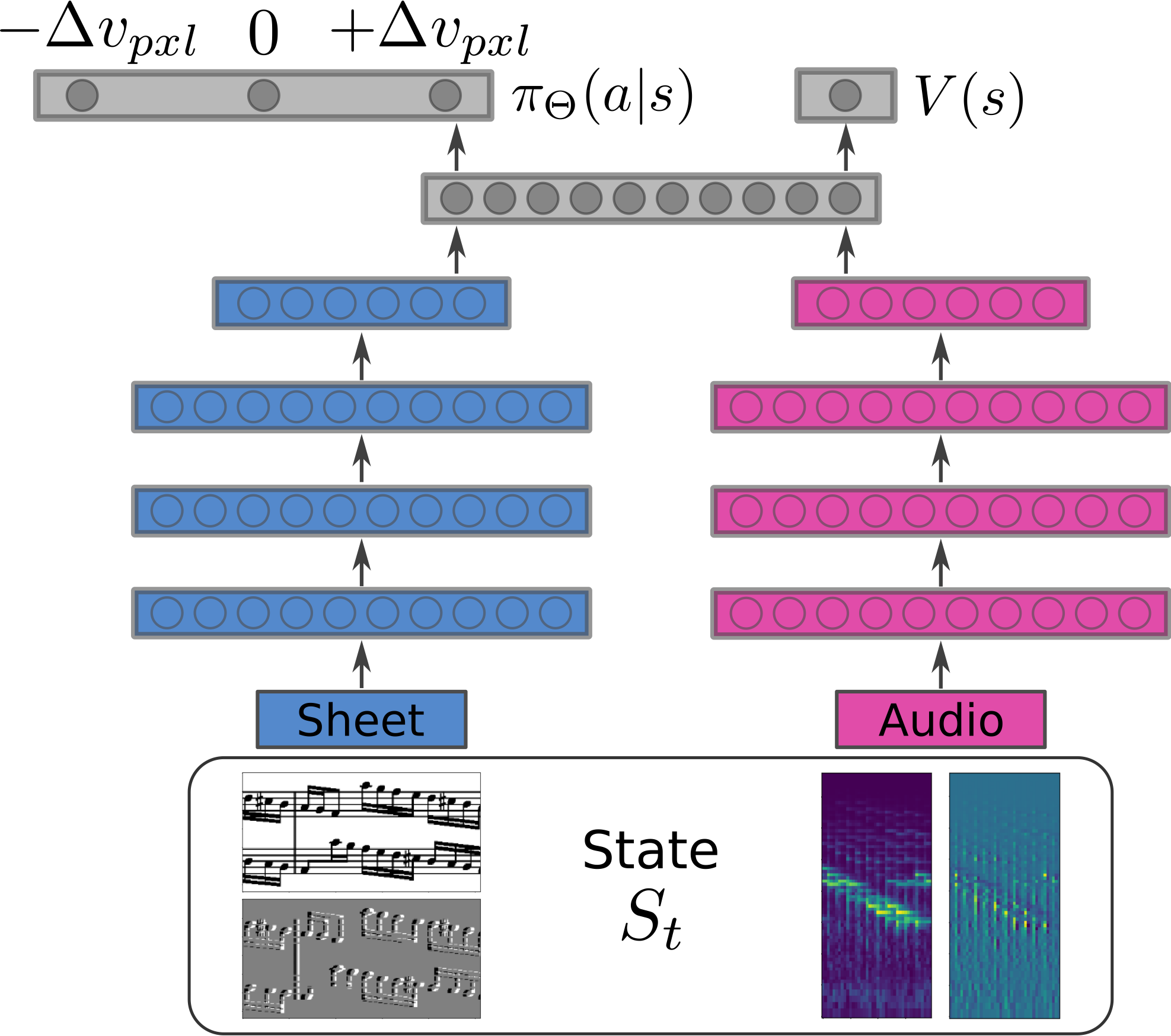}}
 \caption{Multimodal network architecture used for our score following agents.
 Given state $s$ the policy network predicts the action selection probability $\pi_{\Theta}(a|s)$
 for the allowed action $A_t \in \{-\Delta v_{pxl}, 0, +\Delta v_{pxl}\}$.
 The value network, sharing parameters with the policy network,
 provides a state-value estimate $V(s)$ for the current state.
}
\label{fig:network}
\end{figure}

Figure \ref{fig:network} shows a sketch of such a network architecture.
As the authors in \cite{Dorfer2016Towards}, we use a multimodal convolutional neural network
operating on both sheet music and audio at the same time.
The input to the network is exactly the Markov state of the MDP introduced in Section \ref{subsec:states}.
The left part of the network processes sheet images,
the right part spectrogram excerpts (including $\Delta$ images).
After low-level representation learning,
the two modalities are merged by concatenation
and further processed using dense layers.
This architecture implies that policy and value network share the parameters of the lower layers,
which is a common choice in RL \cite{Mnih_2016_A3C}.
Finally, there are two output layers:
the first represents our policy and predicts the action selection probability $\pi_{\Theta}(a|s)$.
It contains three output neurons (one for each possible action)
converted into a valid probability distribution via soft-max activation.
The second output layer consists of one linear output neuron predicting the value $V(s)$ of the current state.
Table \ref{tab:rl_model_architecture} lists
the exact architectures used for our experiments.
We use exponential linear units for all but the two output layers \cite{clevert2015ELU}.
\begin{table}[t]
\small
\caption{\small Network architecture. DO: Dropout, Conv($3$, stride-1)-$16$: 3$\times$3 convolution, 16 feature maps and stride 1.}
\vspace*{2mm}
\scriptsize
\centering
\begin{tabular}{c|c}
\hline
Audio (Spectrogram) $78 \times 40$ & Sheet-Image $80 \times 256$ \\
\hline
Conv($3$, stride-1)-$32$				&	Conv($5$, stride-(1, 2))-$32$ \\
Conv($3$, stride-1)-$32$				&	Conv($3$, stride-1)-$32$ \\
Conv($3$, stride-2)-$64$				&	Conv($3$, stride-2)-$64$ \\
Conv($3$, stride-1)-$64$ + DO(0.2)		&	Conv($3$, stride-1)-$64$ + DO(0.2) \\
Conv($3$, stride-2)-$64$				&	Conv($3$, stride-2)-$64$ \\
Conv($3$, stride-2)-$96$				&	Conv($3$, stride-2)-$64$ + DO(0.2) \\
Conv($3$, stride-1)-$96$				&	Conv($3$, stride-2)-$96$ \\
Conv($1$, stride-1)-$96$ + DO(0.2)		&	Conv($1$, stride-1)-$96$ + DO(0.2)\\
Dense(512)								&	Dense(512) \\
\hline
\multicolumn{2}{c}{Concatenation + Dense(512)} \\
\hline
Dense(256) + DO(0.2)					&	Dense(512) + DO(0.2) \\
Dense(3) - Softmax						& 	Dense(1) - Linear \\
\hline
\end{tabular}
\label{tab:rl_model_architecture}
\end{table}

\subsection{Learning a Policy via Actor-Critic}
One of the first algorithms proposed for optimizing a policy was REINFORCE \cite{Williams_1992_Reinforce},
a Monte-Carlo algorithm that learns by generating entire episodes $S_0, A_0, R_1, S_1, A_1, R_2, S_2, A_2, ...$
of states, actions and rewards by following policy $\pi_{\Theta}$ while interacting with the environment.
Given this sequence it updates the parameters $\Theta$ of the policy network
according to the following update rule
by replaying the episode time step by time step:
\begin{equation}
\Theta \leftarrow \Theta + \alpha G_t \nabla_{\Theta} \ln \pi_{\Theta}(A_t|S_t, \Theta)
\label{eq:reinforce}
\end{equation}
$\alpha$ is the step size or learning rate and
$G_t$ is the true discounted cumulative future reward (the return) received from time step $t$ onwards.
Gradient $\nabla_{\Theta}$ is the direction in parameter space in which to go
if we want to maximize the selection probability of the respective action.
This means whenever the agent did well,
achieving a high return $G_t$,
we take larger steps in parameter space
towards selecting the responsible actions.
By changing the parameters of the policy network,
we of course also change our policy (behavior)
and we will select beneficial actions more frequently in the future
when confronted with similar states.

REINFORCE and policy optimization are known to
have high variance in the gradient estimate \cite{Greensmith_2004_VarianceReduction}.
This results in slow learning and poor convergence properties.
To address this problem,
\textbf{REINFORCE with Baseline} ($\text{REINFORCE}_{bl}$) adapts the update rule of Equation (\ref{eq:reinforce})
by subtracting the estimated state value $V(s)$ (see Section \ref{subsec:reward}) from the actual return $G_t$ received:
\begin{equation}
\Theta \leftarrow \Theta + \alpha (G_t - V(s)) \nabla_{\Theta} \ln \pi_{\Theta}(A_t|S_t, \Theta)
\label{eq:reinforce_bl}
\end{equation}
This simple adaptation helps to reduce variance and improves convergence.
The value network itself is learned by minimizing the mean squared error between the
actually received return and the predicted value estimate of the network, $(G_t - V(s))^2$.
$\text{REINFORCE}_{bl}$ will be the first learning algorithm considered in our experiments.

\emph{Actor-critic methods} are an extension of the baseline concept, allowing agents
to learn in an online fashion while interacting with the environment.
This avoids the need for creating entire episodes prior to learning.
In particular, our actor-critic agent will only look into the future a fixed number of $t_{max}$ time steps (in our case, 15).
This implies that we do not have the actual return $G_t$ available for updating the value function.
The solution is to \emph{bootstrap} the value function (i.e., update the value estimate with estimated values),
which is the core characteristic of actor-critic methods.
The authors in \cite{Mnih_2016_A3C} propose the \textbf{Synchronous Advantage Actor Critic (A2C)}
and show that running multiple actors (in our case 16)
in parallel on different instances of the same kind of environment,
further helps to stabilize training.
We will see in our experiments that this also holds for the score following task.
For a detailed description of the learning process
we refer to the original paper \cite{Mnih_2016_A3C}.


%% file: experiments.tex
In this section we experimentally evaluate our RL approach to score following
and compare it to a previously introduced method \cite{Dorfer2016Towards} that solves the same task.
In addition to quantitative analysis we also provide a video of 
our agents interacting with the score following environment.\footnote{score following video: \url{https://youtu.be/COPNciY510g}}

%
\subsection{Experimental Setup}
Two different datasets will be used in our experiments.
The \emph{Nottingham Dataset} comprises 296 monophonic melodies of folk music (training: 187, validation: 63, testing: 46);
it was already used in \cite{Dorfer2016Towards} to evaluate score following in sheet music images.
The second dataset contains 479 classical pieces by various composers such as Beethoven, Mozart and Bach,
collected from the freely available \emph{Mutopia Project}\footnote{\url{http://www.mutopiaproject.org/}}
(training: 360, validation: 19, testing: 100).
It covers polyphonic music and is a substantially harder challenge to a score follower.
In both cases the sheet music is typeset with Lilypond
and the audios are synthesized from MIDI using an acoustic piano sound font.
This automatic rendering process provides the precise audio -- sheet music alignments required for training (see Section \ref{sec:data}).
For audio processing we set the computation rate to 20 FPS and
compute log-frequency spectrograms at a sample rate of 22.05kHz.
The FFT is computed with a window size of 2048 samples
and post-processed with a logarithmic filterbank allowing only
frequencies from 60Hz to 6kHz (78 frequency bins).

The spectrogram context visible to the agents is set to 40 frames (2 sec. of audio)
and the sliding window sheet images cover $160 \times 512$ pixels
and are further downscaled by a factor of two before being presented to the network.
As optimizer we use the \emph{Adam} update rule \cite{Kingma2014adam}
with an initial learning rate of $10^{-4}$
and running average coefficients of $0.5$ and $0.999$.
We then train the models until there is no improvement in the number of tracked onsets
on the validation set for $50$ epochs
and reduce the learning rate by factor $10$ three times.
The tempo change action $\Delta v_{pxl}$ is $0.5$ for Nottingham and $1.0$ for the polyphonic pieces.

%
\subsection{Evaluation Measures and Baselines}
Recall from Section \ref{subsec:reward} and Figure~\ref{fig:reward}
that from the agent's position $\hat{x}$
and the ground truth position $x$, we compute the tracking error $d_x$.
This error is the basis for our evaluation measures.
However, compared to training,
we only consider time steps in our evaluation where there is actually an onset present in the audio.
While interpolating intermediate time steps is helpful for
creating a stronger learning signal (Section \ref{subsec:reward}), it is not musically meaningful.
Specifically, we will report the evaluation statistics
\emph{mean absolute tracking error} $\overline{|d_x|}$
as well as its standard deviation $std(|d_x|)$ over all test pieces.
These two measures quantify the \emph{accuracy} of the score followers.
To also measure their \emph{robustness} we compute the ratio $R_{on}$ of overall tracked onsets
as well as the ratio of pieces $R_{tue}$ tracked from beginning entirely to the end.

As \emph{baseline method} we consider the approach described in \cite{Dorfer2016Towards},
which models score following as a multimodal localization task
(denoted by \baseline in the following).

\begin{figure}[t!]
 \centerline{\includegraphics[width=0.96\columnwidth]{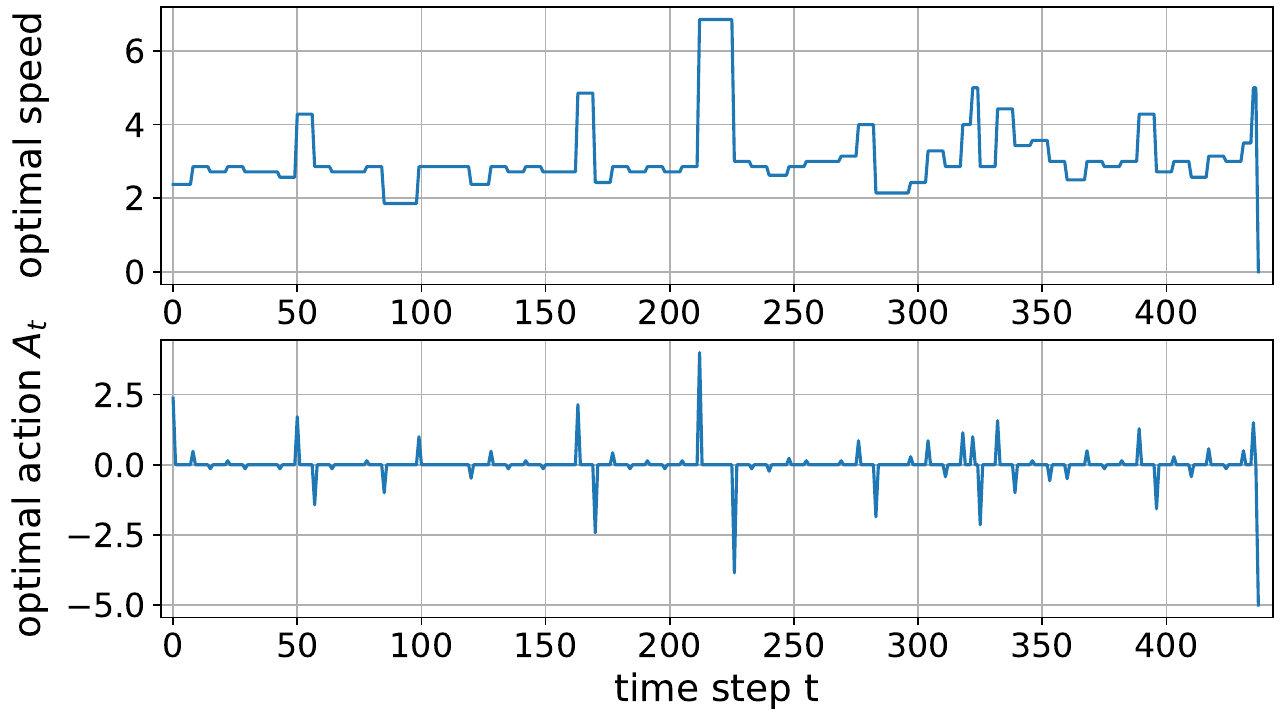}}
 \caption{Optimal tempo curve and corresponding optimal actions $A_t$ for a continuous agent (piece: J. S. Bach, BWV994).
  The $A_t$ would be the target values for training an agent with supervised, feed-forward regression.
}
\label{fig:tempo_curve}
\end{figure}

As a \emph{second baseline}, we also tried to train an agent
to solve the score following MDP in a fully supervised fashion.
This is theoretically possible, as we know for each time point
the exact corresponding position in the score image,
which permits us to derive an optimal tempo curve and, consequently,
an optimal sequence of tempo changes for each of the training pieces.
Figure \ref{fig:tempo_curve} shows such an optimal tempo curve
along with the respective tempo change actions for a short Bach piece.
The latter would serve as targets $y$
in a supervised regression problem $y=f(x)$.
The network structure used for this experiment
is identical to the one in Figure~\ref{fig:network} except for the output layers.
Instead of policy $\pi_{\theta}$ and value $V$ we only keep a single linear output neuron
predicting the value of the optimal tempo change in each time step.
However, a closer look at Figure \ref{fig:tempo_curve} already reveals the problem
inherent in this approach.
The optimal tempo change is close to zero most of the time.
For the remaining time steps we observe sparse spikes of varying amplitude.
When trying to learn to approximate these optimal tempo changes (with a mean squared error optimization target),
we ended up with a network that predicts values very close to zero for all its inputs.
We conclude that the relevant tempo change events are too sparse
for supervised learning and exclude the method from our tables in the following.
Besides these technical difficulties we will also discuss
conceptual advantages of addressing score following as an MDP
in Section \ref{sec:conclusion}.

%
\subsection{Experimental Results}
Table \ref{tab:eval_methods} provides a summary of the experimental results.
%
\begin{table}[t!]
 \small
 \begin{center}
 \begin{tabular}{rcccc}
 \toprule
 \textbf{Method} & $R_{tue}$ & $R_{on}$ & $\overline{|d_x|}$ & $std(|d_x|)$ \\
 \midrule
 \\
 & \multicolumn{4}{c}{Nottingham (monophonic, 46 test pieces)} \\
 \midrule
 \baseline \cite{Dorfer2016Towards}	& 0.43 & 0.65 & 3.15 & 13.15	\\
 \midrule
 $\text{REINFORCE}_{bl}$ & 0.94 & 0.96 & 4.21 & 4.59 \\
 A2C			& \textbf{0.96} & \textbf{0.99} & \textbf{2.17} & \textbf{3.53}  \\
 \bottomrule
 \\
 & \multicolumn{4}{c}{Mutopia (polyphonic, 100 test pieces)} \\
 \midrule
 \baseline \cite{Dorfer2016Towards}	& 0.61 & 0.72 & 62.34 & 298.14 \\
 \midrule
 $\text{REINFORCE}_{bl}$ & 0.20 & 0.35 & 48.61 & 41.99 \\
 A2C					 & \textbf{0.74} & \textbf{0.75} & \textbf{19.25} & \textbf{23.23} \\	
 \bottomrule
 \end{tabular}
\end{center}
 \caption{Comparison of score following approaches. Best results are marked in bold.
 For A2C and $\text{REINFORCE}_{bl}$ we report the average over 10 evaluation runs.}
 \label{tab:eval_methods}
\end{table}
Looking at the Nottingham dataset,
we observe large gaps in performance
between the different approaches.
Both RL based methods manage to follow almost
all of the test pieces completely to the end.
In addition, the mean tracking error is lower for A2C and shows a substantially lower standard deviation.
The high standard deviation for \baseline is even more
evident in the polyphonic pieces.
The reason is that \baseline is formulated as a localization task,
predicting a location probability distribution over the score image given the current audio.
Musical passages can be highly repetitive, which
leads to multiple modes in the location probability distribution, each of which is equally probable.
As the \baseline tracker follows the mode with highest probability
it starts to jump between such ambiguous structures, producing
a high standard deviation for the tracking error
and, in the worst case, loses the target.

Our MDP formulation of score following addresses this issue,
as the agent controls its progression speed for navigating through the sheet image.
This restricts the agent as it does not allow for large jumps in the score
and, in addition, is much closer to how music is actually performed
(e.g. from left to right and top to bottom when excluding repetitions).
Our results (especially the ones of A2C) reflect this theoretical advantage.

However, in the case of complex polyphonic scores
we also observe that the performance of $\text{REINFORCE}_{bl}$ degrades completely.
The numbers reported are the outcome of more than five days of training.
We already mentioned in Section \ref{sec:rl} that policy optimization
is known to have high variance in the gradient estimate \cite{Greensmith_2004_VarianceReduction},
which is exactly what we observe in our experiments.
Even though $\text{REINFORCE}_{bl}$ managed to learn a useful policy for the
Nottingham dataset it also took more than five days to arrive at that.
In contrast, A2C learns a successful policy for the Nottingham dataset in less than six hours
and outperforms the baseline method on both datasets.
For Mutopia it tracks more than 70\% of the 100 test pieces entirely to the end
without losing the target a single time.
This result comes with an average error of only 20 pixels
which is about $5$mm in a standard A4 page of Western sheet music -- three times more accurate than the baseline with a mean error of 62 pixels.

We also report the results of $\text{REINFORCE}_{bl}$ to emphasize the potential of RL in this setting.
Recall that the underlying MDP is the same for both $\text{REINFORCE}_{bl}$ and A2C.
The only part that changes is a more powerful learner.
All other components including network architecture, optimization algorithm
and environment remain untouched.
Considering that deep RL is currently one of the most intensively researched areas
in machine learning, we can expect further improvement in the score following task
whenever there is an advance in RL itself.

%% file: conclusion.tex
We have proposed a formulation of score following in sheet music images as a Markov decision process
and showed how to address it with state-of-the-art deep reinforcement learning.
Experimental results on monophonic and polyphonic piano music show that this is competitive
with recently introduced methods\cite{Dorfer2016Towards}.
We would like to close with a discussion of some specific aspects that point to interesting future perspectives.

Firstly, we trained all agents using a continuous reward signal
computed by interpolating the target (\emph{ground truth}) location between successive onsets and note heads.
Reinforcement learners can, of course, also learn from a \emph{delayed} signal
(e.g. non-zero rewards only at actual onsets or even bar lines or downbeats).
%
This further implies that we could, for example,
take one of our models trained on the synthesized audios,
annotate a set of real performance audios at the bar level (which is perfectly feasible),
and then fine-tune the models with the very same algorithms, with the
sole difference that for time points without annotation
the environment simply returns a neutral reward of zero.

Secondly, we have already started to experiment with continuous control agents that directly predict the required tempo changes,
rather than relying on a discrete set of action.
Continuous control has proven to be very successful in other domains \cite{Duan_2016_RLControl}
and would allow for a perfect alignment of sheet music and audio (cf. Figure \ref{fig:tempo_curve}).

A final remark concerns RL in general.
For many RL benchmarks we are given a simulated environment
that the agents interact with.
These environments are fixed problems
without a natural split into training, validation and testing situations.
This is different in our setting,
and one of the main challenges is to learn agents,
which generalize to unseen pieces and audio conditions.
While techniques such as weight-decay, dropout \cite{Srivastava_2014_Dropout}
or batch-normalization \cite{Ioffe_2015_Batchnorm} have become a standard tool for regularization
in supervised learning they are not researched in the context of RL.
A broad benchmark of these regularizers in the context of RL
would be therefore of high relevance.

We think that all of this makes the score following MDP
a promising and in our opinion very exciting playground for further research
in both music information retrieval and reinforcement learning.